\documentclass[journal]{IEEEtran}

\usepackage{mathtools} 
\usepackage{graphicx}
\usepackage{array}
\usepackage{xcolor}
\usepackage{gensymb}

\newcommand{\red}[1]{\textcolor{black}{#1}}

\begin{document}

\title{Vine Robots: Design, Teleoperation, and Deployment for Navigation and Exploration}

\author{By Margaret M. Coad, Laura H. Blumenschein, Sadie Cutler, Javier A. Reyna Zepeda, Nicholas D. Naclerio,\\ Haitham El-Hussieny, Usman Mehmood, Jee-Hwan Ryu, Elliot W. Hawkes, and Allison M. Okamura}

\maketitle

\IEEEPARstart{A}{}new class of \red{continuum} robots has recently been explored, characterized by tip extension, significant length change, and directional control. Here, we call this class of robots ``vine robots," due to their similar behavior to plants with the growth habit of trailing. Due to their growth-based movement, vine robots are well suited for navigation and exploration in cluttered environments, but until now, they have not been deployed outside the lab. Portability of these robots and steerability at length scales relevant for navigation are key to field applications. In addition, intuitive human-in-the-loop teleoperation enables movement in unknown and dynamic environments. We present a vine robot system that is teleoperated using a custom designed flexible joystick and camera system, long enough for use in navigation tasks, and portable for use in the field. We report on deployment of this system in two scenarios: a soft robot navigation competition and exploration of an archaeological site. The competition course required movement over uneven terrain, past unstable obstacles, and through a small aperture. The archaeological site required movement over rocks and through horizontal and vertical turns. The robot tip successfully moved past the obstacles and through the tunnels, demonstrating the capability of vine robots to achieve navigation and exploration tasks \red{in the field}.

\section*{Background}
There are a number of potential applications of robotics where non-destructive exploration of small spaces remains challenging for existing robot design. These include inspection~\cite{liu2013state}, search and rescue~\cite{murphy2016disaster}, medicine~\cite{taylor2016medical}, and archaeology~\cite{khatib2016ocean}. Vine robots can potentially fill this need for robots that can move in highly constrained environments.
 
\red{Unlike continuum robots that lengthen by extending modules that move relative to the environment until fully emitted~\cite{wooten2015novel, neumann2016considerations}, vine robots emit new material only at the very tip, which enables lengthening without relative movement between the emitted robot body material and the environment. Extension in this way also enables enormous length change, limited only by the amount of material that can be transported to the tip. This form of extension has been realized by different mechanisms~\cite{sadeghi2017toward}, but here} we focus on vine robots that lengthen using internal air pressure to pass the material of their flexible, tubular body through its center and turn it inside out at the tip, through a process called eversion. Mishima et al.~\cite{mishima2003development} list three benefits of pneumatically everting vine robots for navigation in cluttered environments: flexibility to follow tortuous paths, rigidity to support their own weight while traversing gaps, and ability to access spaces without movement of their body relative to the environment. Tsukagoshi et al.~\cite{tsukagoshi2011tip} point out two additional benefits of pneumatically everting vine robots: their bodies could be used as conduits to deliver water or other payloads to their tip, and since their movement is driven completely by air pressure rather than electricity, they do not risk igniting flammable gases in hazardous environments due to failure of electrical components. Hawkes et al.~\cite{HawkesScienceRobotics2017} demonstrate the ability of pneumatically everting vine robots to move with ease through sticky, slippery, and sharp environments and to grow into useful structures through preformed shapes. \red{We hypothesize that the unique features of vine robots allow execution of navigation and exploration tasks in ways not achievable by other types of robots.} Various proof of concept designs have been developed for pneumatically~\cite{mishima2003development, tsukagoshi2011tip, HawkesScienceRobotics2017, greer2018soft}\red{, and hydraulically~\cite{luong2019eversion}} everting vine robot navigation and exploration systems. Building on these proof of concept designs, there is a need for a complete vine robot system suitable for deployment \red{in the field for} navigation and exploration scenarios. This article presents such a system.

The contributions of this work are:
\begin{itemize}
\item A complete, portable system for vine robot deployment in the field. \red{Our vine robot system combines the capabilities of the proof of concept designs and is steerable, carries a camera, and grows to an arbitrary length from a compact form factor.}
\item A reversible steering vine robot actuator that is easily manufactured at long lengths. \red{We improve upon the design of the actuator presented in~\cite{greer2018soft, GreerICRA2017} by creating a body-length steering actuator that can be manufactured by heat sealing and attached to the robot body using double-sided tape.}
\item A method of mounting a camera at the tip of a vine robot and managing the camera wire using a rigid cap and zipper pocket. \red{In contrast to the bulky design presented in~\cite{mishima2003development}, the limited-length design presented in~\cite{tsukagoshi2011tip, HawkesScienceRobotics2017, greer2018soft}, and the wireless design presented in~\cite{luong2019eversion}, we contribute a compact tip camera mount and wire management design that allows vine robot growth from a compact base to an arbitrary length.}
\item A method for \red{robust} control of vine robot growth speed using a motor to restrict growth. \red{In contrast to~\cite{luong2019eversion}, our controller prevents the lack of control that would occur if the motor's speed were faster than the pressure-driven growth speed.}
\item A geometric model-based method for teleoperated steering of vine robots using a custom-designed flexible joystick. \red{We adapt the teleoperation device presented in~\cite{ElHussienyIROS2018} and the mapping from desired tip position to actuator pressure  presented in~\cite{greer2018soft, GreerICRA2017} to achieve human-in-the-loop teleoperation.}
\item A report on vine robot deployment experience in two different locations: a soft robot navigation competition and an archaeological site. \red{Moving beyond the demonstrations of vine robots completing navigation tasks in laboratory environments presented in~\cite{tsukagoshi2011tip, HawkesScienceRobotics2017, greer2018soft}, we deploy vine robots in the field.}
\end{itemize}

\section*{Vine Robot System Requirements}  
A number of design requirements were considered for the design of our vine robot system, stemming from basic vine robot functionality, our goal of making a system capable of operating in unpredictable environments, and the specific deployment scenarios, which tested the vine robot's ability to achieve navigation and exploration tasks \red{in the field}. \red{The following subsections present the details of the specific scenarios, as well as a summary of the system's design requirements.}

\subsection*{Soft Robot Navigation Competition \red{Scenario}}

The first deployment opportunity for our vine robot system was the soft robot navigation competition at the IEEE International Conference on Soft Robotics in Livorno, Italy in April 2018 (RoboSoft 2018). The competition was designed to test capabilities considered fundamental for soft robots, such as ``mechanical compliance, delicate interaction with the environment, and dexterity"~\cite{calisti2016contest}, and provided a way to benchmark the capabilities of different robots. The competition course was based on a mock disaster scenario, where a robot would enter a building and navigate challenging terrain both inside and outside the building. The 9.5 m-long course consisted of four obstacles: a sand pit, a square aperture, stairs, and a set of unstable cylinders that were easily knocked over. 
The competition had a task completion-based scoring system. The sand pit and the stairs only needed to be crossed to achieve full points. The aperture's size was chosen by each team, with smaller apertures relative to the robot's diameter yielding higher points when traversed. The unstable cylinders needed to be passed through without knocking any of them over to achieve full points.

\subsection*{Archaeological Exploration \red{Scenario}}
The second deployment opportunity for our vine robot system was for exploration of an archaeological site in Chavin, Peru in July 2018. The archaeological site was a monumental center of religion and culture for the ancient Andean civilization that flourished there between approximately 1200 and 500 BC~\cite{kembel2004building}, and parts of the structure remain intact today. Many of the spaces in the site are too small for a human to crawl into and too tortuous to be explored with a camera on a stick, so we were invited to use our vine robot to help explore and take video inside tight spaces at the site. The site contains hundreds of largely unexplored underground tunnels that can range in size from approximately 30 to 100 cm across and stretch up to hundreds of meters long. Exploration of these tunnels is important to the archaeology team because they might lead to other underground rooms in which objects of interest can be found, or they may themselves contain objects of interest. Additionally, mapping the shape of the tunnels might lead to understanding about their purpose or significance to the people who made them. Video and photos of the site, as well as discussions with the archaeology team, helped develop design requirements for the version of the vine robot deployed at the archaeological site.
\red{\subsection*{Summary of Design Requirements}}
The basic requirement for pneumatically everting vine robot design is that the soft robot body be made of a non-stretchable material that is flexible enough to be turned inside out at the tip and that is capable of containing pressurized air. 

Our system is also teleoperated to allow the human operator to make decisions about how to proceed with navigation and exploration in unstructured environments. \red{To achieve effective teleoperation, the robot's growth and steering must be controllable, and there must be an interface for the human operator to give control inputs. There must also be a way for the human operator to observe the position of the robot tip within its environment.}

\red{Additionally, our system must be capable of navigation and exploration tasks, which means it must be able to grow to a length useful for navigation, pass through small apertures, and support its own body weight when navigating vertically and over gaps in the floor of the environment.}

\red{Finally, our system must be usable in the field, which means that it must be portable, mechanically and electrically robust enough to last through the lifetime of use (a single competition run or a week of testing at the archaeological site), able to move fast enough to be practically useful, and, for the case of the archaeological application, able to record data taken during exploration.}

These design requirements are summarized in Table~\ref{requirements}, along with the design solutions chosen for the two slightly different versions of the vine robot deployed in the two locations, \red{which will be explained in the following sections.}

{\renewcommand{\arraystretch}{1.5}
\begin{table*}
\caption{Design Requirements and Solutions for Vine Robot System}
\label{requirements}
\centering
\begin{tabular}{|>{\centering\arraybackslash}p{3.2cm}|>{\centering\arraybackslash}p{4cm}|>{\centering\arraybackslash}p{4.7cm}|>{\centering\arraybackslash}p{4.7cm}|}
\hline
\textbf{Capability} & \textbf{Requirement} & \textbf{Solution for Competition} & \textbf{Solution for Archaeology}\\
\hline
\hline
\textit{Pneumatically Everting Vine Robot} & Flexible, not stretchable, airtight soft robot body material & Thin, airtight plastic & Thin, airtight fabric \\
\hline
\hline
& Controllable growth & \multicolumn{2}{c|}{Antagonistic growth control with motor and pressure regulator} \\
\cline{2-4}
\textit{Teleoperation} & Controllable steering & \multicolumn{2}{c|}{Series pouch motor actuators controlled by pressure regulators} \\
\cline{2-4}
& Human operator control interface & \multicolumn{2}{c|}{Flexible joystick}\\
\cline{2-4}
& Human operator situational awareness & Line of sight to robot tip & Camera at robot tip and display with real-time video feedback \\
\hline
\hline
& Length & \multicolumn{2}{c|}{Soft robot body material stored on reel in robot base} \\
\cline{2-4}
\textit{Navigation/Exploration} & \red{Aperture navigation} & Natural soft robot body shrinking; ~~~~~~~~~~ \red{aperture} width $<$ body diameter & Smooth, rounded camera cap; ~~~~~~~~~~~~~ \red{aperture} width = body diameter \\
\cline{2-4}
& Body support & \multicolumn{2}{c|}{Lightweight body material}\\
\hline
\hline
& Portability & \multicolumn{2}{c|}{Small soft robot body diameter to ensure portable air compressor can fill it} \\
\cline{2-4}
\textit{Usability \red{in the Field}} & Mechanical robustness & \red{Plastic soft robot body} & Durable fabric soft robot body\\
\cline{2-4}
& Electrical robustness & \multicolumn{2}{c|}{All electronics run off line power and all signals are wired}\\
\cline{2-4}
& Speed of movement & \multicolumn{2}{c|}{Fast pressure control, backdrivable motor}\\
\cline{2-4}
& Data recording & n/a & Camera at robot tip with video recording\\
\hline
\end{tabular}\\
\end{table*}

\section*{System Overview}
We present here the design and control of a vine robot system that is teleoperated using visual feedback from a camera at the tip, portable, and not inherently limited in length. Figure~\ref{fig:systemDrawing} shows diagrams of the three main features of the robot system: growth to an arbitrary length, reversible steering using soft pneumatic actuators, and transport of a camera at the robot tip. Figure~\ref{fig:systemPic} shows the complete system, made up of the growing portion of the robot, the base station, and the human interface for controlling the robot. The following sections discuss in detail each component of the system. \red{Table~\ref{specs} lists the design specifications for the two different versions of the robot, which will be explained as the appropriate components are discussed.}
  
  \begin{figure}
  \centering
  \includegraphics[width=.9\columnwidth]{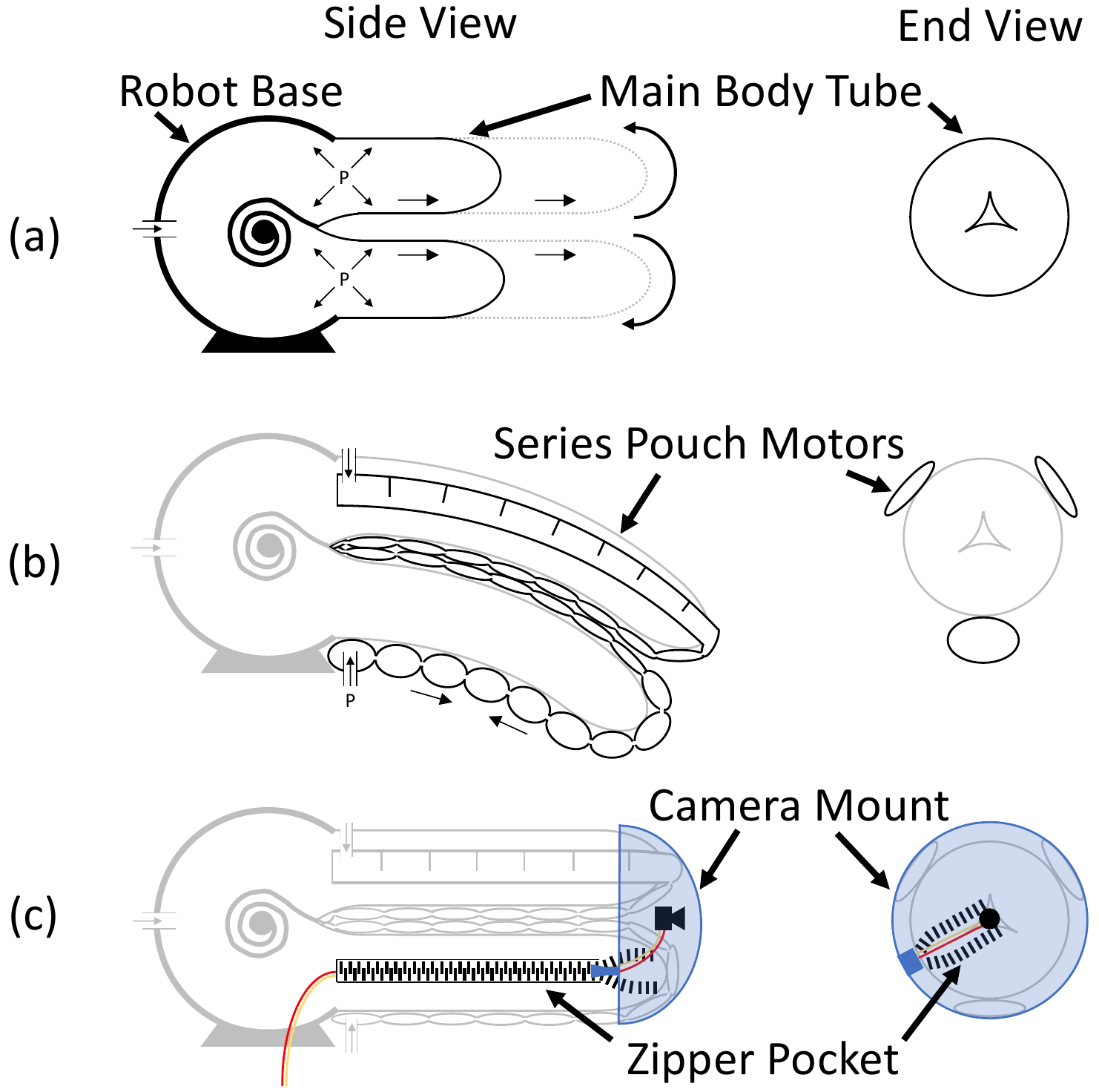}
  \caption{
Three main features of our vine robot system: (a) growth to an arbitrary length enabled by storing soft robot body material on a spool inside the robot base, (b) reversible steering of the robot tip using series pouch motor soft pneumatic actuators that run the entire length of the robot body, and (c) transport of a wired camera at the robot tip using a rigid cap that is pushed along as the robot body grows, as well as a pocket to contain the camera wires that runs the entire length of the robot body and is zipped up as the robot body grows.}
  \label{fig:systemDrawing}
  \end{figure}
  
 \begin{figure}
  \centering
  \includegraphics[width=\columnwidth]{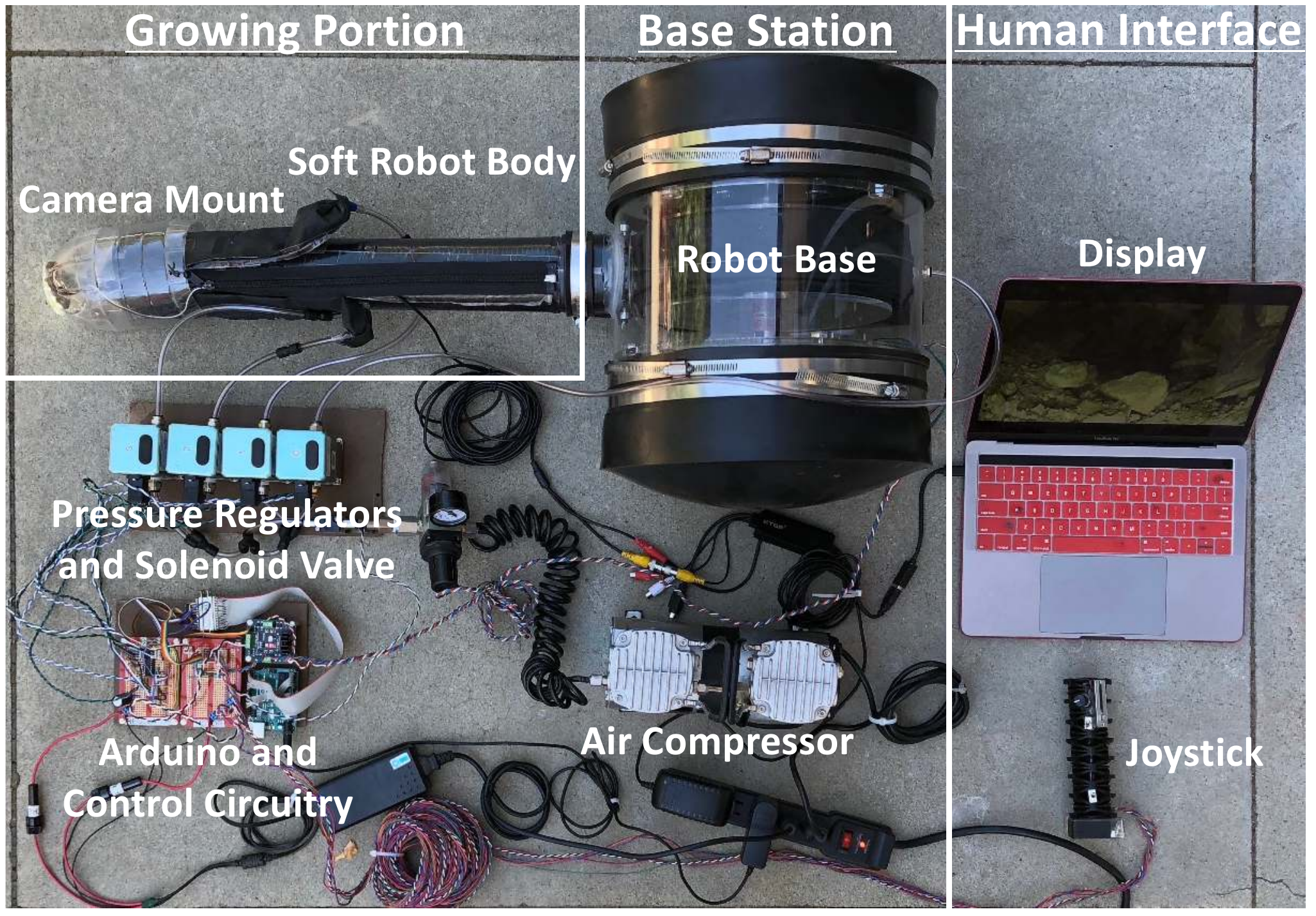}
  \caption{Complete vine robot system. Components include the growing portion (which contains the soft robot body and camera mount), base station (which contains the mechanical, electrical, and pneumatic components required to move the growing portion), and the human interface (which contains the flexible joystick and the display for viewing camera images).}
  \label{fig:systemPic}
  \end{figure}
  
\begin{table}
\caption{Design Specifications for Vine Robot System}
\label{specs}
\centering
\begin{tabular}{|>{\centering\arraybackslash}p{3.4cm}|>{\centering\arraybackslash}p{2cm}|>{\centering\arraybackslash}p{2cm}|}
\hline
\textbf{Feature} & \textbf{Specification for Competition} & \textbf{Specification for Archaeology}\\
\hline
\hline
\textit{Soft robot body material} & 0.005 cm thick LDPE$^*$ & 0.015 cm thick TPU-nylon$^{**}$ \\
\hline
\textit{Soft robot body length} & 10 m & 7.5 m \\
\hline
\textit{Main body tube diameter} & 5 cm & 7.5 cm \\
\hline
\textit{Actuator tube diameter} & 2.5 cm & 3.7 cm \\
\hline
\textit{Robot base diameter} & 20 cm & 30 cm \\
\hline
\textit{Robot base length} & \multicolumn{2}{c|}{30 cm} \\
\hline
\textit{Max growth speed} & \multicolumn{2}{c|}{10 cm/s} \\
\hline
\textit{Max growth pressure} & 14 kPa & 21 kPa \\
\hline
\textit{Max steering pressure} & 14 kPa & 21 kPa \\
\hline
\textit{Max air compressor flow rate} & \multicolumn{2}{c|}{470 cubic cm/s} \\
\hline
\end{tabular}\\
\vspace{5 pt}
\footnotesize{$^*$Low-density polyethylene (Uline, Pleasant Prairie, WI)\\ $^{**}$Thermoplastic polyurethane-coated ripstop nylon (Seattle Fabrics, Inc., Seattle, WA)}\\
\end{table}}

\section*{Mechanical Design}

\subsection*{Soft Robot Body Design}
\begin{figure}
  \centering
  \includegraphics[width=\columnwidth]{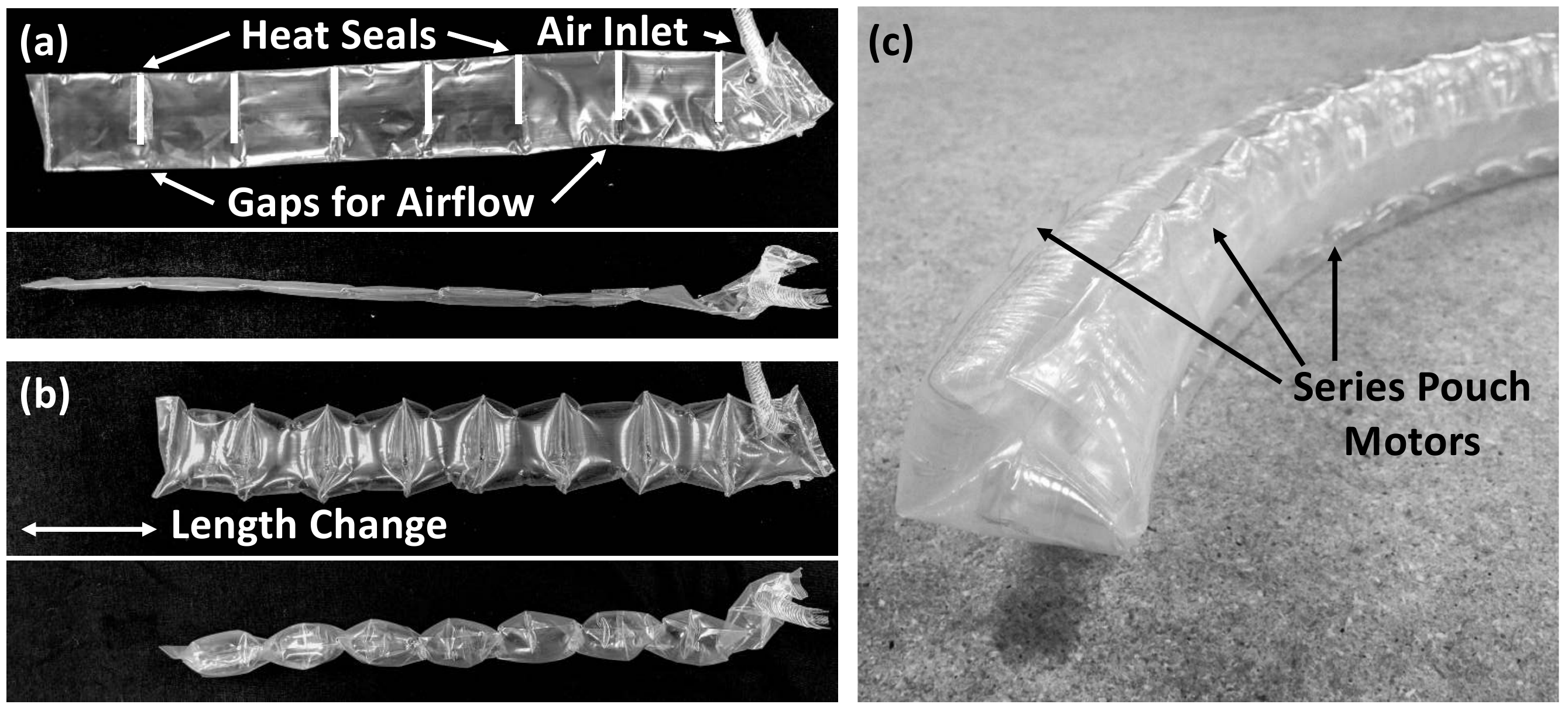}
  \caption{Soft actuation for steering of the robot. (a) Top and side view of the uninflated actuator, constructed by partially heat sealing a tube of airtight, flexible material at regular intervals. (b) Top and side view of the inflated actuator, which balloons out at each pouch, causing shortening along the entire length. (c) Close-up of vine robot tip, showing three series pouch motors spaced equally around the main body tube. One series pouch motor is inflated, causing the robot body to reversibly curve towards it.}
  \label{fig:actuators}
  \end{figure}

The soft body of the vine robot is made of four airtight tubes that are flexible but not stretchable: one central main body tube and three smaller actuator tubes that are placed around the main body tube (Figures~\ref{fig:systemDrawing}(a),~\ref{fig:systemDrawing}(b), and~\ref{fig:actuators}(c)). Growth is achieved by pressurizing the main body tube. One end of the main body tube is fixed to an opening in a rigid pressure vessel (Figure~\ref{fig:base}), and the other end of the tube is folded inside of itself and wrapped around a spool inside the pressure vessel. This allows a long length of robot body material to be stored in a compact space. Pressurizing the pressure vessel, and thus the main body tube, \red{while allowing the body material to unroll from the spool}, causes the robot body to elongate from the tip. \red{Thin, airtight plastic was chosen for the competition robot body, because it could be purchased in a tube shape, which allowed rapid prototyping and manufacturing. Thin, airtight fabric was chosen for the archaeological exploration robot body, because a more durable material was needed to withstand repeated use in the abrasive environment of the tunnels. Both materials are lightweight to allow the robot body to support its own weight. The soft robot body length was chosen to be just long enough to complete the competition course or to achieve useful exploration at the archaeological site. The soft robot body diameter was chosen to be large enough to allow growth at a low pressure~\cite{blumenschein2017modeling} but small enough for the air compressor to quickly fill the robot body's increasing volume during growth. The diameter of the archaeology robot was slightly larger than that of the competition robot, because the additional thickness of the fabric meant that a larger diameter was needed to grow at the same pressure.}

Reversible steering of the robot body is achieved using the three actuator tubes, each of which is partially heat sealed at regular intervals to create a series pouch motor soft pneumatic actuator that shortens when inflated, as shown in Figure~\ref{fig:actuators}. These actuators are based on the pouch motors presented in~\cite{niiyama2015pouch} and are arranged in series like the series pneumatic artificial muscles presented in~\cite{greer2018soft, GreerICRA2017}, yielding a design that is easily manufactured at long lengths \red{and easily attached to the main body tube. We chose the actuator tube diameters to be as large as possible while allowing a small gap between neighboring actuators.} The three series pouch motors are attached lengthwise to the exterior of the main body using double-sided tape (MD 9000, Marker-Tape, Mico, TX), equally spaced around the circumference of the main body tube. When one of the series pouch motors is pressurized, the length change that it produces causes the entire robot to curve in the direction of that actuator (Figures~\ref{fig:systemDrawing}(b) and~\ref{fig:actuators}(c)). The three shortening actuators move the robot tip in two degrees of freedom on a \red{surface} in 3D space, and the third degree of freedom of robot tip motion is produced through growth. Growth and steering can occur simultaneously.

\subsection*{Base Station Design}
\begin{figure}
  \centering
  \includegraphics[width=\columnwidth]{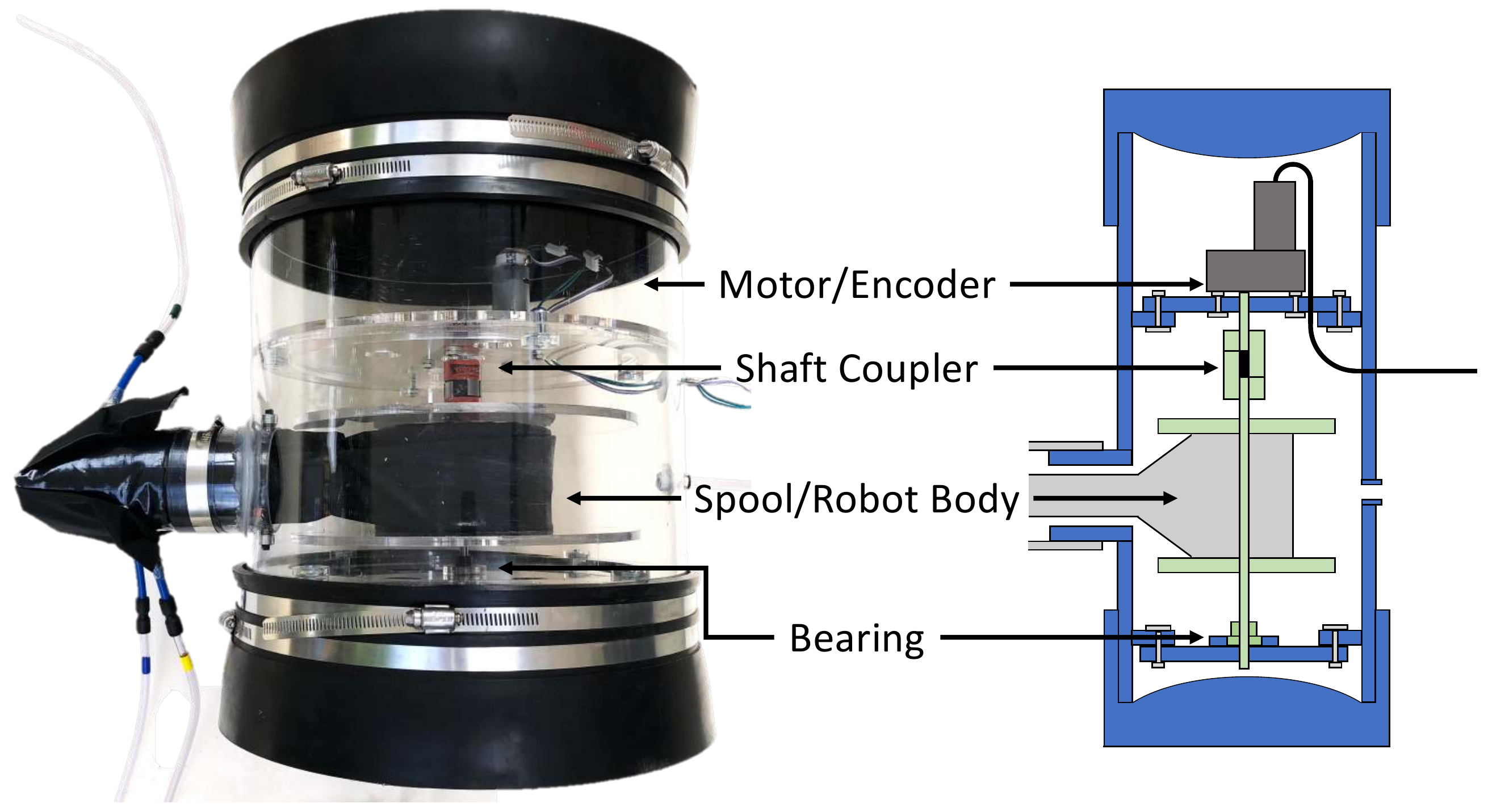}
  \caption{The base used to grow and retract the robot. Robot body material is stored inside a pressurized cylinder on a spool driven by a motor/encoder to control the speed of growth and aid in retraction.}
  \label{fig:base}
  \end{figure}

Control of the vine robot body's motion is enabled by the mechanical, electrical, and pneumatic components of the base station. The robot base (Figure~\ref{fig:base}), a cylindrical pressure vessel made by enclosing a large acrylic cylinder with two end caps (QC-108 or QC-112, Fernco, Inc., Davison, MI), is used to store the undeployed robot body material on a spool. A second, smaller cylinder is fixed inside a hole in the large cylinder using hot glue, and the base of the main body tube of the vine robot is clamped to this smaller cylinder to create an airtight seal. In order to allow the robot body to grow to full length and still be pulled back after deployment, the distal end of the main body tube is attached to a string the length of the robot body that is tied to the spool in the base. The spool is driven by a motor (CHM-2445-1M, Molon, Arlington Heights, IL) with an encoder (3081, Pololu Corporation, Las Vegas, NV), which allows controlled release of the robot body material during growth and assists with retraction of the robot body material back into the base. \red{The length of the robot base was chosen to contain the motor and spool assembly, and the diameter of the base was chosen to contain the rolled up soft robot body. The base for the archaeological exploration needed to be larger in diameter than the base for the competition so as to store the thicker soft robot body material.}

In addition to the robot base, the base station includes pressure regulators, control circuitry, an air compressor, and a solenoid valve. Control of the air pressure in the four tubes of the robot body is achieved using four closed-loop pressure regulators (QB3TANKKZP10PSG, Proportion-Air, Inc., McCordsville, IN), shown in Figure~\ref{fig:systemPic}. An Arduino Uno (Arduino, Turin, Italy), signal conditioning circuitry, and a motor driver (DRI0002, DFRobot, Shanghai, China) control the voltages sent to the motor and pressure regulators. A portable air compressor (FS-MA1000B, Silentaire Technology, Houston, TX) provides a continuous supply of compressed air to the system. \red{For the competition, the provided air compressor was used, which has the same maximum flow rate and also has a storage tank.} A fail-closed solenoid valve (MME-31NES-D012, Clippard, Cincinnati, OH) sits in-line between the air compressor and the pressure regulators to allow quick release of all pressure in the system in case of emergency or power failure. \red{One determining factor of the vine robot's maximum growth speed is the maximum flow rate of compressed air through the system, so high flow-rate pneumatic components were selected. The air compressor ended up being the component that limited the system's overall maximum flow rate. For robustness, all connections in the system are wired, and no part of the system runs on battery power. This allows continuous operation in the field, provided that power lines are available.}

\subsection*{Flexible Joystick Design}
\begin{figure}
  \centering
  \includegraphics[width=\columnwidth]{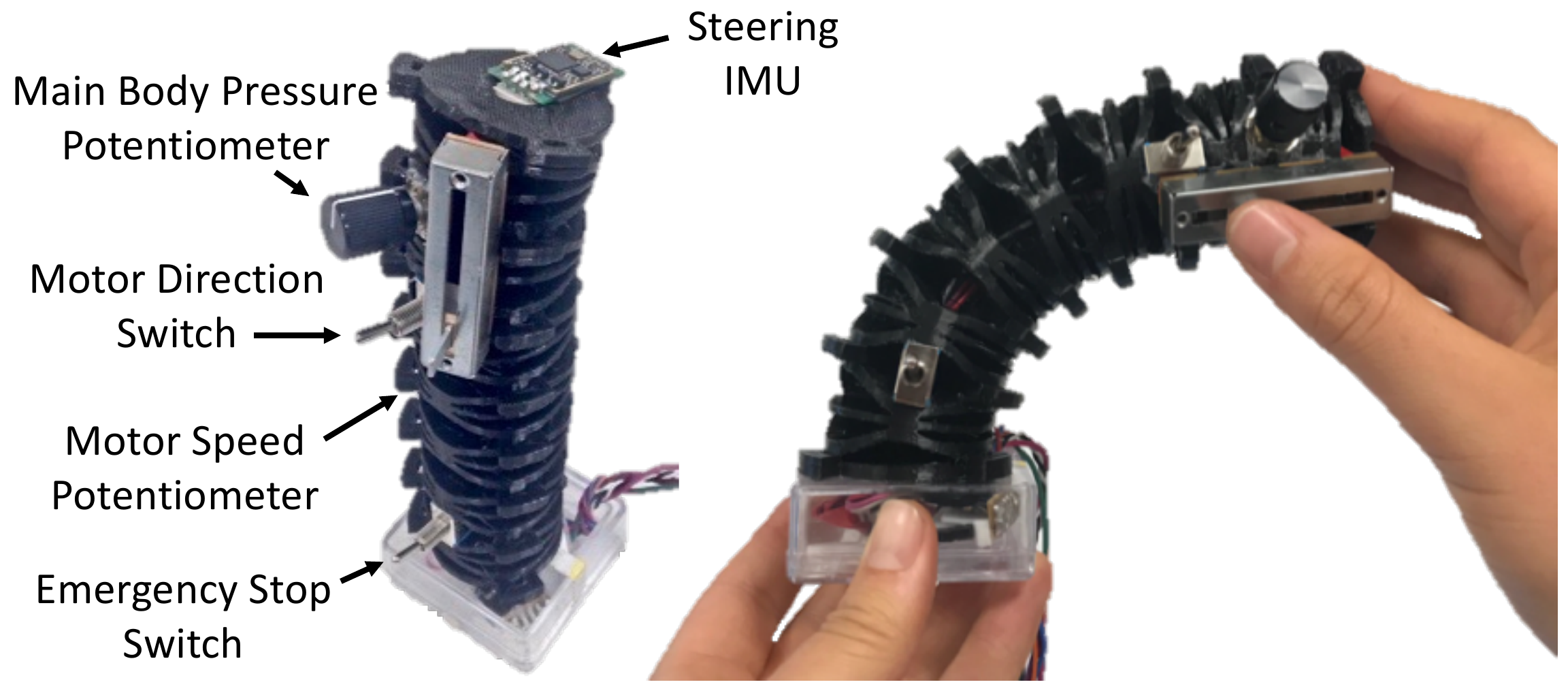}
  \caption{The flexible joystick used for teleoperation. The joystick contains switches and sensors to control the motion of the soft robot body.}
  \label{fig:joystick}
  \end{figure}

We use a flexible joystick, first presented in~\cite{ElHussienyIROS2018} and adapted with additional control switches and potentiometers, as the interface for a human operator to teleoperate the vine robot. The mechanical design and components of the joystick are shown in Figure~\ref{fig:joystick}. The shape of the joystick mimics the long, thin, bendable shape of the soft body of the vine robot, allowing the human operator to steer it in an intuitive way~\cite{ElHussienyIROS2018}. The joystick is made out of 3D printed flexible rubber (NinjaFlex, NinjaTek, Manheim, PA). An inertial measurement unit (IMU) (EBIMU-9DOFV3, E2BOX, Hanam, South Korea) on the flexible joystick measures the orientation of the tip. 

\begin{figure}
  \centering
  \includegraphics[width=\columnwidth]{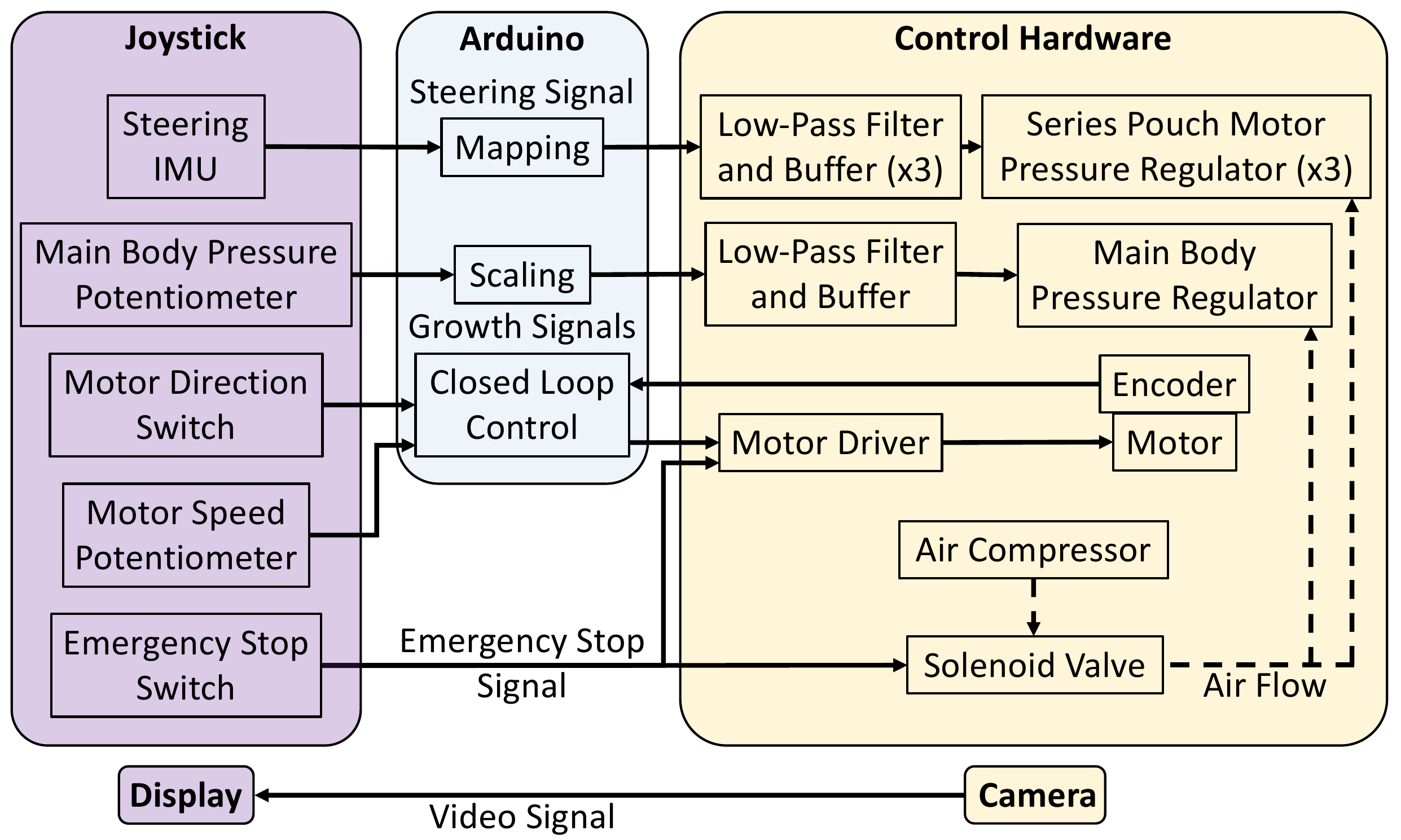}
  \caption{A diagram of the vine robot's electrical signals and air flow. Components of the human interface are shown in purple, the Arduino is shown in blue, and components of the sensing and control hardware are shown in yellow. Electrical signals are shown with a solid line, and the flow of compressed air is shown with a dotted line.}
  \label{fig:electronics}
  \end{figure}

In addition to the IMU, which controls the steering of the vine robot, the flexible joystick also contains inputs for main body pressure and motor speed, which together control the growth (Figure~\ref{fig:electronics}). A rotary potentiometer sets the pressure in the main body, and a sliding potentiometer sets the desired motor speed. The joystick also includes an emergency-stop toggle switch.

\subsection*{Camera Mount Design}
\begin{figure}
  \centering
  \includegraphics[width=\columnwidth]{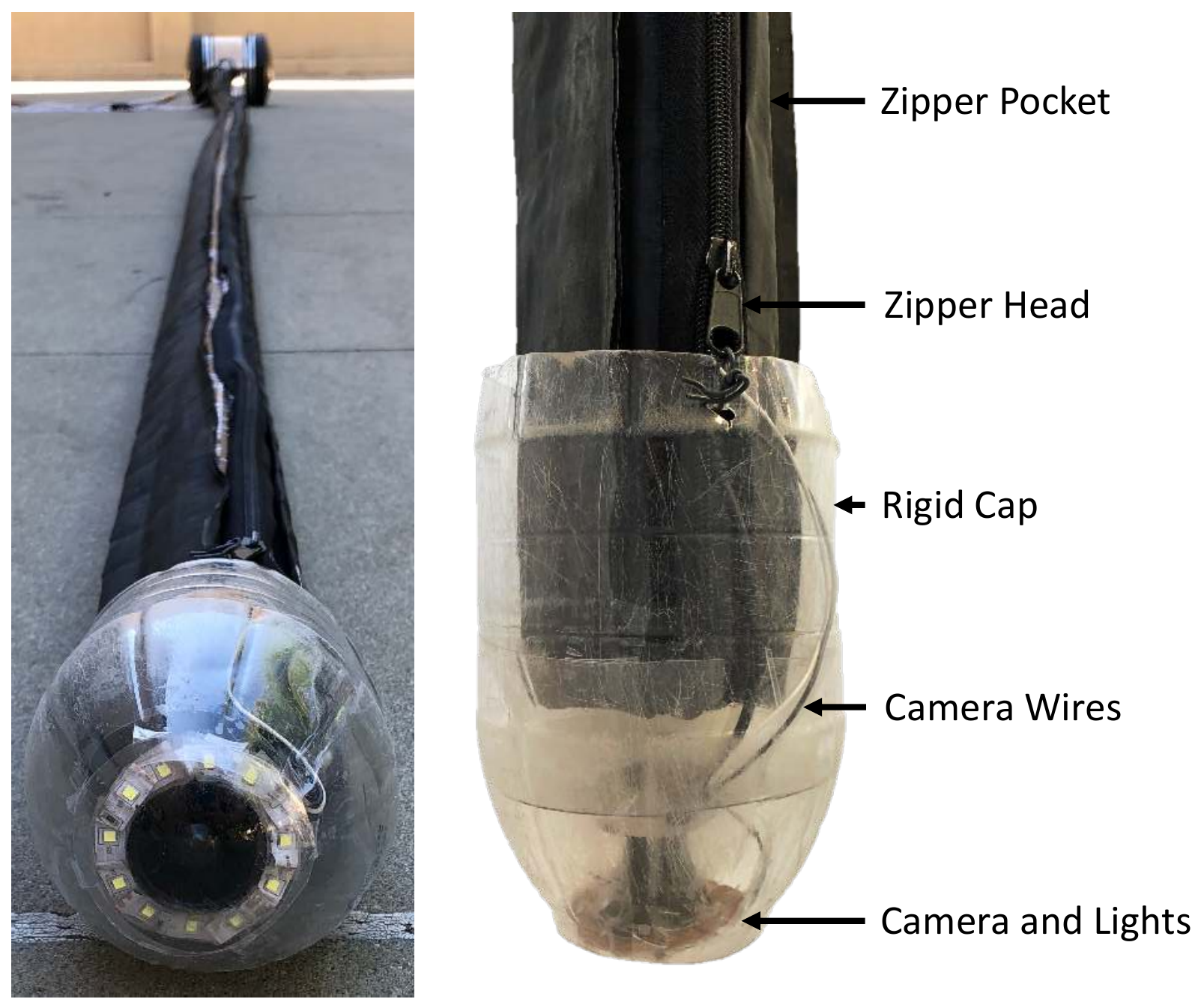}
  \caption{The camera mount system. The camera and lights are contained in a clear, rigid cap that gets pushed along as the vine robot grows. The camera wires are stored at the robot base and slide through a zipper pocket that grows with the robot.}
  \label{fig:cameraCap}
  \end{figure}

\red{For the competition, the focus was on navigation of the robot tip through the obstacles, and the operator was allowed direct line of sight of the robot tip, so no camera was needed. However, for the archaeological exploration, the focus was on exploration and data collection in an unknown environment, so} we mounted a camera (HDE-S62-NEW, SpyCamPro, Toronto, Canada) at the robot tip to allow teleoperation and video recording. The camera is mounted to a rigid cap, which stays at the robot tip during growth as shown in Figure~\ref{fig:cameraCap}. The cap's inner diameter (10.4 cm) is slightly larger than the outer diameter of the soft robot body when one or two of the actuator tubes is inflated, which allows the cap to slide freely along the robot body and be pushed along by the robot tip as the robot grows. The bullet shape guides the cap to slide along walls or obstacles that it contacts from any angle. A strip of LED lights surrounds the camera to illuminate the environment in front of the robot. The camera has a wide angle lens to allow it to capture approximately 120\degree~of the environment without moving.

Since wireless signals are difficult to transmit underground, we used a wired camera for archaeological exploration. The camera wire is stored coiled up at the base of the robot, and as the soft robot body grows, the wire is pulled along the exterior of the robot body by the camera cap. In order to prevent the wire from snagging on the environment, an LDPE pocket on the soft robot body contains the camera wire and allows it to slide inside the pocket without directly contacting the environment. The pocket lengthens as the soft robot body grows using a zipper mechanism (Figures~\ref{fig:systemDrawing}(c) and~\ref{fig:cameraCap}). 
A zipper runs the entire length of the pocket, with the base of the zipper at the base of the robot. 
The zipper head is fixed to the camera cap so that the zipper starts out unzipped when the soft robot body is short and zips up as the robot grows, thus creating a pocket that is always the length of the soft robot body. The zipper also prevents rotation of the camera cap relative to the soft robot body, simplifying the mapping between camera image movement and actuator movement.

\section*{Control}
Figure~\ref{fig:electronics} shows the flow of information between the human interface and the sensing and control hardware of our vine robot system. This section describes in detail the mapping, scaling, and closed-loop control performed within the Arduino to convert joystick inputs into the motor voltage and the four pressures used to control the movement of the robot body. Growth and steering are controlled independently and occur simultaneously. Since we do not sense the robot shape, our controller relies on the human operator to close the loop via line of sight or camera feedback to achieve a desired robot tip position.

\subsection*{Growth Control}
Growth control is enacted by balancing the main body pressure with the motor voltage. The main body pressure is directly set using the main body pressure potentiometer as
\begin{eqnarray}\label{pressure_pot}
p = c_p(r_p-r_{p0}),
\end{eqnarray}
where $p$ is the desired pressure in the main body, $c_p$ is a constant that converts units of potentiometer readings to units of pressure, $r_p$ is the current potentiometer reading, and $r_{p0}$ is the potentiometer reading at the position that corresponds to zero pressure. A closed-loop pressure regulator runs its own internal control loop to maintain a desired pressure given an analog voltage input. The Arduino PWM signal is sent through a low-pass filter and buffer to create a true analog voltage input for the pressure regulators.

The desired motor speed $\omega_d$ is commanded with the motor direction switch and motor speed potentiometer as
\begin{eqnarray}\label{motor_pot}
\omega_d = d~c_m(r_m-r_{m0}),
\end{eqnarray}
where $d$ equals $-1$ if the motor direction switch is in the growth direction and $1$ if the motor direction switch is in the retraction direction, $c_m$ is a constant that converts units of potentiometer readings to units of motor speed, $r_m$ is the current potentiometer reading, and $r_{m0}$ is the potentiometer reading at the position that corresponds to zero motor speed.
The desired motor speed is maintained using a proportional-integral control loop based on readings from the encoder attached to the motor, and the motor voltage control signal $u$ is calculated as
\begin{eqnarray}\label{motor_pi}
u = k_p(\omega_d - \omega)+k_i\int (\omega_d - \omega),
\end{eqnarray}
where $k_p$ is the proportional control constant, $k_i$ is the integral control constant, and $\omega$ is the actual motor speed as measured by the encoder.

Because only pressure can cause the robot to grow and only motor voltage can cause the robot to retract, there is a delicate balance between pressure and motor voltage to allow controllability of growth. For smooth growth to occur, the main body pressure must be higher than the pressure needed to grow~\cite{blumenschein2017modeling,NaclerioIROS2018}, and the motor must maintain tension in the robot body and/or string coming off the spool. If the motor spins faster in the growth direction than the robot is growing, the robot material or string will become slack, and the human operator will lose control over slowing the growth. For this reason, \red{we use a backdrivable motor to restrain the robot's growth.} In our teleoperation controller, if the calculated motor voltage control signal would cause rotation and/or torque of the motor in the growth direction, the motor voltage is instead set to just cancel the Coulomb friction in the gearing of the motor. \red{This allows the motor to be easily backdriven by the string or robot body and to unspool material when needed while never unspooling material too quickly. The maximum growth pressure used was 14 kPa for the competition robot and 21 kPa for the archaeology robot, and the maximum observed growth speed for both systems was approximately 10 cm/s.}

\subsection*{Steering Control}

Steering control is achieved by using the measured orientation of the IMU at the tip of the joystick to determine the desired position of the soft robot body tip within a shell defined by the two degrees of freedom of movement not governed by growth~\cite{greer2018soft}. Movement of the robot tip to this position is then enacted in an open-loop fashion by setting the desired pressures of the three closed loop pressure regulators that supply air to the three series pouch motor actuators. 

First, the IMU-measured joystick tip orientation, $\textbf{\textit{q}}$, represented in quaternion form, is used to calculate the curvature amount $\kappa$ and the direction of curvature (i.e. bending plane angle) $\phi$ of the joystick. Based on the constant curvature model of continuum robots~\cite{webster2010design}, these shape parameters are calculated as 
\begin{eqnarray}\label{kappaphi}
\begin{aligned}
&\kappa =\dfrac{\cos^{-1}\left(1 - 2 (q^2_{x} +q^2_{y})\right)}{s}, \ \kappa>0,\\
&\phi =\tan^{-1}\left(\dfrac{q_{x} q_{w} + q_{y} q_{z}}{q_{x} q_{z}- q_{y} q_{w}}\right),\  -\pi\leq\phi< \pi,
\end{aligned}
\end{eqnarray}
where $q_{w}$ and $[q_{x},q_{y},q_{z}]^T$ are the scalar and the vector components of  $\textbf{\textit{q}}$, and $s$ is the length of the flexible joystick. Then, the $x$ and $y$ coordinates of the 3D position of the joystick tip relative to its base are calculated using
\begin{eqnarray}\label{xy_joystick}
\begin{aligned}
&x_{\text{joystick}} =-\dfrac{\cos(\phi)(\cos(\kappa s)-1)}{\kappa},\\ 
&y_{\text{joystick}} =\dfrac{\sin(\phi)(\cos(\kappa s)-1)}{\kappa}.
\end{aligned}
\end{eqnarray}

Next, the desired robot tip coordinates are set equal to the current joystick tip coordinates, and the movement of the robot tip to these coordinates is enacted through setting the three series pouch motor pressures based on a simple geometric model of the soft robot body \red{adapted from the geometric and static model presented in~\cite{greer2018soft, GreerICRA2017}}. Pressurization of each series pouch motor is assumed to cause movement of the robot tip towards that series pouch motor with a displacement proportional to the pressure. The resulting position of the robot tip is assumed to be a superposition of the displacements produced by each series pouch motor, as
\begin{eqnarray}\label{xy}
\begin{aligned}
&x = c(p_1\cos(\psi_1)+p_2\cos(\psi_2)+p_3\cos(\psi_3)),\\ 
&y = c(p_1\sin(\psi_1)+p_2\sin(\psi_2)+p_3\sin(\psi_3)),
\end{aligned}
\end{eqnarray}
where $c$ is a tunable constant that converts units of pressure into units of robot tip displacement and controls the amount of curvature enacted in the soft robot body for a given movement of the joystick, $p_1$, $p_2$, and $p_3$ are the pressures sent to the three series pouch motors, and $\psi_1$, $\psi_2$, and $\psi_3$ are the angles counterclockwise from the positive $x$ axis at which the three series pouch motors are placed around the circumference of the soft robot body. Due to the weight of the soft robot body, only the curvature of its most distal 1 meter (approximately) can be controlled by the human operator, while the rest of its body tends to remain fixed. This results in approximately the same robot tip movement at various robot body lengths. The pressures sent to the actuators are calculated by solving Equations (\ref{xy}) as
\begin{eqnarray}\label{p}
\begin{aligned}
&p_1 = \dfrac{\sin(\psi_3-\psi_2)}{\sin(\psi_2-\psi_1)}p_3 + \dfrac{x\sin(\psi_2)-y\cos(\psi_2)}{d\sin(\psi_2-\psi_1)}, \ \\ 
&p_2 = \dfrac{\sin(\psi_3-\psi_1)}{\sin(\psi_1-\psi_2)}p_3 + \dfrac{x\sin(\psi_1)-y\cos(\psi_1)}{d\sin(\psi_1-\psi_2)}, \ \\ 
&p_3 = p_3.
\end{aligned}
\end{eqnarray}
Since there are three actuators but only two degrees of freedom of steering, there is a redundancy in the actuation, which we handle by always ensuring that the pressure in at least one of the actuators is close to zero. We start from an initial guess of zero for the value of $p_3$ and calculate $p_1$ and $p_2$. Then, we iteratively update the guess for $p_3$ and re-solve for $p_1$ and $p_2$ until all of the calculated pressures are positive and at least one of the calculated pressures is within a small tolerance of zero. This avoids unnecessary shortening and stiffening of the robot body due to co-contraction of opposing series pouch motors. \red{The maximum steering pressure used was 14 kPa for the competition robot and 21 kPa for the archaeology robot.}

\section*{Deployment at a Soft Robot Navigation Competition}

\begin{figure*}
  \centering
  \includegraphics[width=\textwidth]{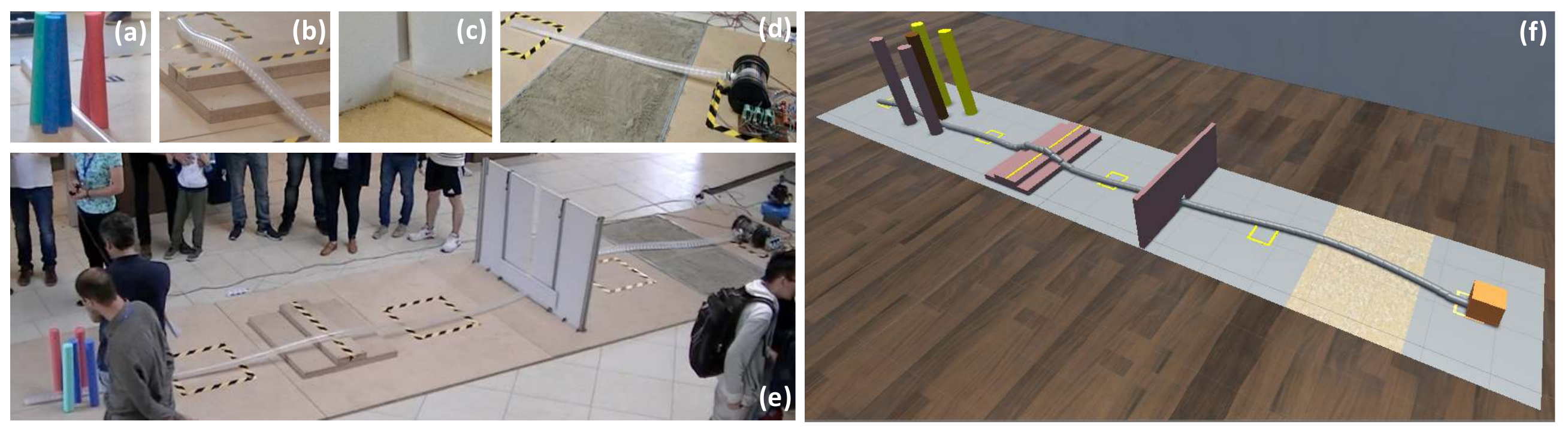}
  \caption{Photos and simulation of the vine robot's successful completion of the RoboSoft 2018 soft robot navigation competition course, consisting of (a) unstable cylinders that were easily knocked over, (b) stairs, (c) a small aperture, and (d) a sand pit. (e) The vine robot after completing the entire competition course. (f) Simulation of the vine robot's execution of the course. The vine robot was the only robot in the competition to navigate all obstacles perfectly on the first attempt, and it also passed through the smallest aperture (4.5 cm square) relative to its body size (7 cm diameter).}
  \label{fig:competition}
  \end{figure*}

Seven robots from around the world competed in the RoboSoft 2018 soft robot navigation competition. Figure~\ref{fig:competition} shows the vine robot successfully executing the four obstacles: the sand pit, the square aperture, the stairs, and the unstable cylinders. Overall, the vine robot was the only robot in the competition to navigate all obstacles perfectly on the first attempt, and it also passed through the smallest aperture \red{overall, as well as the smallest aperture} relative to its body size. However, due to its growth-based movement (robots lose points when they do not move their whole body through each obstacle), the robot received half points on the first three obstacles, which placed it third overall.

During practice, we were consistently able to teleoperate the vine robot through the course in under three minutes (6 cm per second). The sandpit did not present a problem, since the vine robot does not rely on exerting forces on the environment to move straight, \red{like typical ground-locomoting robots do.} \red{Due to the hollow, air-filled inside of the soft robot body,} the robot was consistently able to \red{shrink its diameter while passing} through an aperture with a side length of four centimeters for a robot with a seven centimeter diameter when all four tubes were inflated (yielding a body shrinking ratio of 0.57:1); any smaller and the robot body would buckle and/or slide along the wall instead of going through the aperture. The robot had no trouble traversing the stairs obstacle, since the actuators could provide enough curvature to grow over each step. The robot could pass through the unstable cylinders without knocking them over, due to its low center of gravity and gentle contact, but it always slid one out of the way, due to its inability to make a tight S-shaped curve without the environment holding its body in place.

At the competition, one of the closed loop pressure regulators had broken in transit, so the main body tube pressure was controlled by hand. Additionally, a leak caused in transit required greater than the maximum flow rate of the air compressor (470 cubic centimeters per second). This caused the storage tank on the air compressor to empty three times during the competition run, requiring pausing of growth to wait for the tank to refill. Despite these (correctable) challenges, the robot was able to execute all four obstacles perfectly on the first try. The total time required to complete the course was 13 minutes and 28 seconds.

\section*{Deployment at an Archaeological Site}
\begin{figure*}
  \centering
  \includegraphics[width=\textwidth]{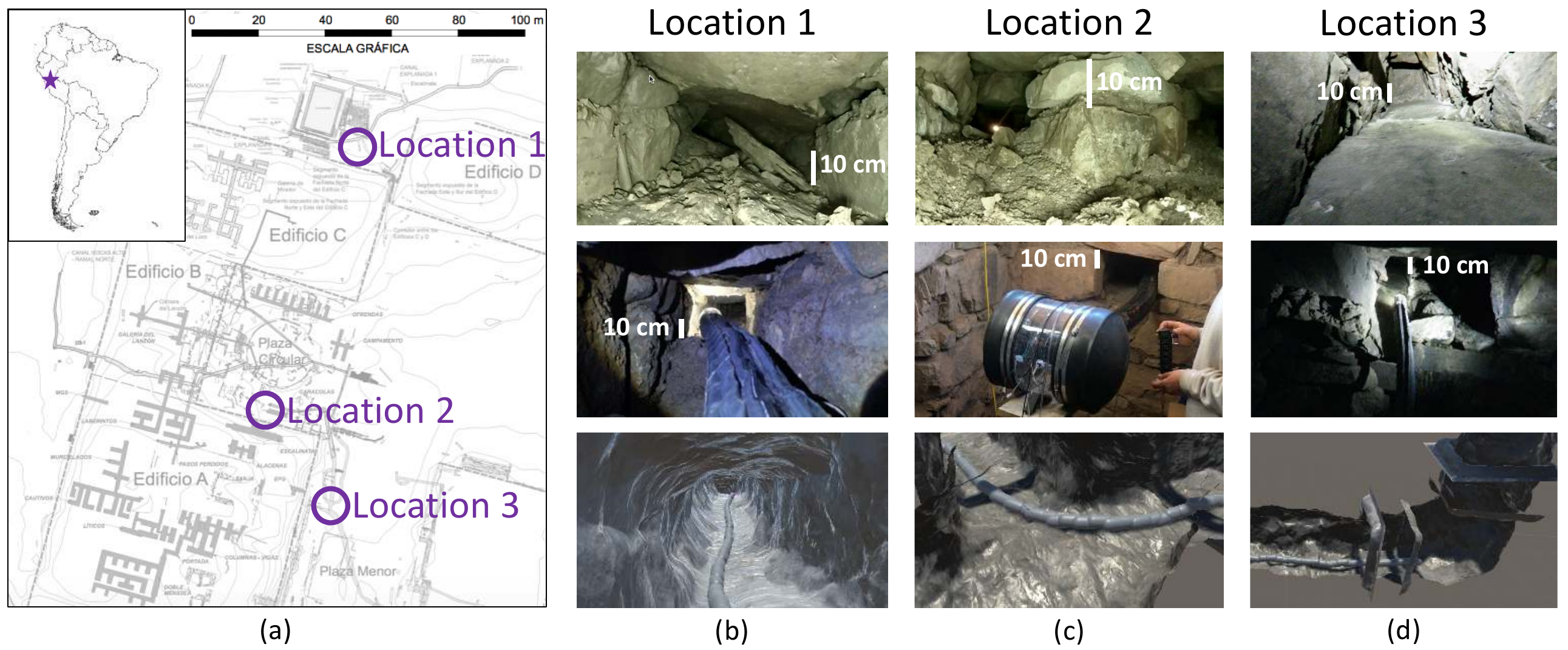}
  \caption{Map, photos, and simulation of the vine robot's successful exploration of underground tunnels in an archaeological site in Chavin, Peru. (a) A map of the archaeological site. Areas the vine robot explored are marked in purple. (b) Location 1: A tunnel almost completely blocked by rocks. (c) Location 2: A tunnel with a 90 degree right hand turn. (d) Location 3: A tunnel that started sloping upwards and then turned completely vertical. Top row are photos from inside the tunnels, middle row are photos of the vine robot in the tunnels, and bottom row are screenshots of the simulation of the vine robot in each tunnel.}
  \label{fig:archaeology}
  \end{figure*}

Figure~\ref{fig:archaeology} shows a map of the archaeological site as well as photos and simulations of the locations explored by the vine robot. Three locations were chosen to be explored by the vine robot due to their interest to the archaeology community, difficulty to explore through other means, expected length (less than 10 meters from a human-sized entry way), and ease of setting up the vine robot at the entrance. Overall, the robot was able to achieve access inside all three of the targeted locations and take video that could not have been recorded otherwise. In Location 1, the robot was able to navigate past a rock blockage (Figure~\ref{fig:archaeology}(b), top). In Location 2, the robot was able to round a 90 degree turn (Figure~\ref{fig:archaeology}(c), top and bottom). In Location 3, the robot was able to grow upwards into a vertical shaft (Figure~\ref{fig:archaeology}(d)). \red{The robot grew approximately 6 m, 5 m, and 3 m into each tunnel, respectively.}

The challenges during this deployment of the robot were artificially slow growth speed, lack of actuator robustness, lack of shape morphing at the robot tip, inability to shorten the robot once grown, and difficulty maintaining situational awareness. First, the speed of growth of the vine robot was artificially reduced due to a significant pressure drop between the closed loop pressure regulators and the soft robot body caused by too long and narrow pressure tubing. Second, the heat seals on the actuators tended to pop open after repeated use, leading to leaks and inability to curve the robot body. This was later improved by stapling over the heat seals and taping over the staples. Third, having a rigid camera cap at the robot tip allowed mounting and protection of the camera, but it also inhibited the vine robot's natural ability to grow along walls and squeeze through narrow apertures. This led to the need to push the robot forward from the base at some points. Fourth, due to the robot's natural tendency to buckle rather than \red{reverse growth} when the motor is run in the retraction direction, it was impossible to \red{retract} the robot while in the tunnels, resulting in the need to pull the robot back from the base to undo wrong turns and remove the robot after deployment. Fifth, challenges with situational awareness came from teleoperating the robot based only on the tip camera image. Because the tip of the robot body sometimes rolled relative to its base, changing the alignment of the camera image with gravity, the mental mapping between the bending directions of the joystick and the world-grounded directions in the tunnels was not always intuitive. Also, it was difficult to maintain an understanding of how far and in what direction the robot tip had gone, leading to confusion about the state of the robot and its environment. Even with these challenges, the vine robot successfully achieved access inside all three tunnels and recorded video in locations not previously observed by the archaeology team.

\section*{Discussion and Future Work}

In this article, we presented a complete vine robot system for use \red{in the field for} navigation and exploration tasks, and we reported on deployment of two slightly different versions of this system to successfully navigate the RoboSoft 2018 soft robot navigation competition course and explore an archaeological site in Chavin, Peru.

In the competition, the vine robot was well suited to robust completion of the competition course due to its ability to move over and around obstacles in a manner different than any other robots in the competition. \red{Tasks that provided challenges for other robots, such as the combination of passing through a small aperture and being large enough to surmount the stairs, were easy for the vine robot.} Its only disadvantage in the competition was the penalty that it incurred in the scoring due to its growth-based movement. This raises the point that there are some situations in which leaving behind part of the robot body is not ideal, and in these cases, vine robots would not be the robot of choice. However, there are many scenarios in which the vine robot structure is advantageous, for example in providing a conduit for fluids and electrical signals.

At the archaeological site, the vine robot's strengths were navigation over rocks, around curves, and up vertical shafts,  \red{all having started from a compact form factor. Navigating the sandy and rocky terrain was easier than it likely would have been for locomoting robots that rely on exerting forces on the environment for movement, and fitting into the small entry ways and navigating the long and tortuous paths of the tunnels were easier than they likely would have been for typical elongating continuum robots. However, due to the lack of direct line of sight to the robot tip, exploration of the archaeological site} was not as easy for the vine robot as completing the competition course. One question that remains open is how best to transport a camera or other sensors via a pneumatically everting vine robot without encumbering the robot's natural ability to morph its shape and grow along or over obstacles at its tip. In addition, we need to develop methods for \red{retracting} the vine robot without buckling. Finally, we wish to improve situational awareness for the human operator of a vine robot in an occluded environment where the operator does not have a direct line of sight to the robot tip.

With the design of robust, field-ready vine robots, we aim to improve the state of the art for robots that can non-destructively explore small spaces. Continued research into burrowing with pneumatically everting vine robots~\cite{NaclerioIROS2018} could open doors to navigation in even more restricted spaces than is currently possible. Additionally, using the vine robot's body as a conduit to pass material through it could take advantage of its unique mechanism of movement through growth.

Another goal of this project is to make the design of vine robots accessible for other researchers and end-users. We created a website (vinerobots.org) with step-by-step instructions for making pneumatically everting vine robots without active steering, and we will add designs and control software for other vine robot versions in the future.

\section*{Acknowledgments}
We thank Marcello Calisti and Jamie Paik for organizing the RoboSoft 2018 competition, as well as John Rick, Jack Lane, Daniel Chan, and the Stanford Global Engineering Program for supporting the vine robot's deployment at the archaeological site. Thanks to Jon Stingel and Tariq Zahroof for their contributions to vine robot base design and teleoperation and to Joey Greer for useful discussions about vine robot growth control. This work was supported in part by the Air Force Office of Scientific Research Grant FA2386-17-1-4658, the National Science Foundation Grant 1637446, and the project ``Toward the Next Generation of Robotic Humanitarian Assistance and Disaster Relief: Fundamental Enabling Technologies (10069072)" funded by the Ministry of Trade, Industry, and Energy of S. Korea.

\bibliographystyle{IEEEtran}
\bibliography{library}

\begin{thebibliography}{10}
\providecommand{\url}[1]{#1}
\csname url@samestyle\endcsname
\providecommand{\newblock}{\relax}
\providecommand{\bibinfo}[2]{#2}
\providecommand{\BIBentrySTDinterwordspacing}{\spaceskip=0pt\relax}
\providecommand{\BIBentryALTinterwordstretchfactor}{4}
\providecommand{\BIBentryALTinterwordspacing}{\spaceskip=\fontdimen2\font plus
\BIBentryALTinterwordstretchfactor\fontdimen3\font minus
  \fontdimen4\font\relax}
\providecommand{\BIBforeignlanguage}[2]{{%
\expandafter\ifx\csname l@#1\endcsname\relax
\typeout{** WARNING: IEEEtran.bst: No hyphenation pattern has been}%
\typeout{** loaded for the language `#1'. Using the pattern for}%
\typeout{** the default language instead.}%
\else
\language=\csname l@#1\endcsname
\fi
#2}}
\providecommand{\BIBdecl}{\relax}
\BIBdecl

\bibitem{liu2013state}
Z.~Liu and Y.~Kleiner, ``State of the art review of inspection technologies for
  condition assessment of water pipes,'' \emph{Measurement}, vol.~46, no.~1,
  pp. 1--15, 2013.

\bibitem{murphy2016disaster}
R.~R. Murphy, S.~Tadokoro, and A.~Kleiner, ``Disaster robotics,'' in
  \emph{Springer Handbook of Robotics}.\hskip 1em plus 0.5em minus 0.4em\relax
  Springer, 2016, pp. 1577--1604.

\bibitem{taylor2016medical}
R.~H. Taylor, A.~Menciassi, G.~Fichtinger, P.~Fiorini, and P.~Dario, ``Medical
  robotics and computer-integrated surgery,'' in \emph{Springer Handbook of
  Robotics}.\hskip 1em plus 0.5em minus 0.4em\relax Springer, 2016, pp.
  1657--1684.

\bibitem{khatib2016ocean}
O.~Khatib, X.~Yeh, G.~Brantner, B.~Soe, B.~Kim, S.~Ganguly, H.~Stuart, S.~Wang,
  M.~Cutkosky, A.~Edsinger \emph{et~al.}, ``Ocean one: A robotic avatar for
  oceanic discovery,'' \emph{IEEE Robotics \& Automation Magazine}, vol.~23,
  no.~4, pp. 20--29, 2016.

\bibitem{wooten2015novel}
M.~B. Wooten and I.~D. Walker, ``A novel vine-like robot for in-orbit
  inspection.''\hskip 1em plus 0.5em minus 0.4em\relax 45th International
  Conference on Environmental Systems, 2015, pp. 1--11.

\bibitem{neumann2016considerations}
M.~Neumann and J.~Burgner-Kahrs, ``Considerations for follow-the-leader motion
  of extensible tendon-driven continuum robots,'' in \emph{2016 IEEE
  International Conference on Robotics and Automation (ICRA)}.\hskip 1em plus
  0.5em minus 0.4em\relax IEEE, 2016, pp. 917--923.

\bibitem{sadeghi2017toward}
A.~Sadeghi, A.~Mondini, and B.~Mazzolai, ``Toward self-growing soft robots
  inspired by plant roots and based on additive manufacturing technologies,''
  \emph{Soft Robotics}, vol.~4, no.~3, pp. 211--223, 2017.

\bibitem{mishima2003development}
D.~Mishima, T.~Aoki, and S.~Hirose, ``Development of pneumatically controlled
  expandable arm for search in the environment with tight access,'' in
  \emph{Field and Service Robotics}.\hskip 1em plus 0.5em minus 0.4em\relax
  Springer, 2003, pp. 509--518.

\bibitem{tsukagoshi2011tip}
H.~Tsukagoshi, N.~Arai, I.~Kiryu, and A.~Kitagawa, ``Tip growing actuator with
  the hose-like structure aiming for inspection on narrow terrain.''
  \emph{International Journal of Automation Technology}, vol.~5, no.~4, pp.
  516--522, 2011.

\bibitem{HawkesScienceRobotics2017}
E.~W. Hawkes, L.~H. Blumenschein, J.~D. Greer, and A.~M. Okamura, ``A soft
  robot that navigates its environment through growth,'' \emph{Science
  Robotics}, vol.~2, no.~8, p. eaan3028, 2017.

\bibitem{greer2018soft}
J.~D. Greer, T.~K. Morimoto, A.~M. Okamura, and E.~W. Hawkes, ``A soft,
  steerable continuum robot that grows via tip extension,'' \emph{Soft
  Robotics}, pp. 95--108, 2018.

\bibitem{luong2019eversion}
J.~Luong, P.~Glick, A.~Ong, M.~S. deVries, S.~Sandin, E.~W. Hawkes, and M.~T.
  Tolley, ``Eversion and retraction of a soft robot towards the exploration of
  coral reefs,'' in \emph{2019 2nd IEEE International Conference on Soft
  Robotics (RoboSoft)}.\hskip 1em plus 0.5em minus 0.4em\relax IEEE, 2019, pp.
  801--807.

\bibitem{GreerICRA2017}
J.~D. Greer, T.~K. Morimoto, A.~M. Okamura, and E.~W. Hawkes, ``Series
  pneumatic artificial muscles ({sPAMs}) and application to a soft continuum
  robot,'' in \emph{IEEE International Conference on Robotics and Automation},
  2017, pp. 5503--5510.

\bibitem{ElHussienyIROS2018}
H.~El-Hussieny, U.~Mehmood, Z.~Mehdi, S.-G. Jeong, M.~Usman, E.~W. Hawkes,
  A.~M. Okamura, and J.-H. Ryu, ``Development and evaluation of an intuitive
  flexible interface for teleoperating soft growing robots,'' in \emph{IEEE/RSJ
  International Conference on Intelligent Robots and Systems}, 2018, pp.
  4995--5002.

\bibitem{calisti2016contest}
M.~Calisti, M.~Cianchetti, M.~Manti, F.~Corucci, and C.~Laschi,
  ``Contest-driven soft-robotics boost: the robosoft grand challenge,''
  \emph{Frontiers in Robotics and AI}, vol.~3, p.~55, 2016.

\bibitem{kembel2004building}
S.~R. Kembel and J.~W. Rick, ``Building authority at {Chavin de Huantar}:
  models of social organization and development in the initial period and early
  horizon,'' \emph{Andean Archaeology}, pp. 51--76, 2004.

\bibitem{blumenschein2017modeling}
L.~H. Blumenschein, A.~M. Okamura, and E.~W. Hawkes, ``Modeling of bioinspired
  apical extension in a soft robot,'' in \emph{Conference on Biomimetic and
  Biohybrid Systems (Living Machines)}.\hskip 1em plus 0.5em minus 0.4em\relax
  Springer, 2017, pp. 522--531.

\bibitem{niiyama2015pouch}
R.~Niiyama, X.~Sun, C.~Sung, B.~An, D.~Rus, and S.~Kim, ``Pouch motors:
  Printable soft actuators integrated with computational design,'' \emph{Soft
  Robotics}, vol.~2, no.~2, pp. 59--70, 2015.

\bibitem{NaclerioIROS2018}
N.~Naclerio, C.~Hubicki, Y.~Aydin, D.~Goldman, and E.~W. Hawkes, ``Soft robotic
  burrowing device with tip-extension and granular fluidization,'' in
  \emph{IEEE/RSJ International Conference on Intelligent Robots and Systems},
  2018, pp. 5918--5923.

\bibitem{webster2010design}
R.~J. Webster~III and B.~A. Jones, ``Design and kinematic modeling of constant
  curvature continuum robots: A review,'' \emph{The International Journal of
  Robotics Research}, vol.~29, no.~13, pp. 1661--1683, 2010.

\end{thebibliography}

\begin{IEEEbiographynophoto}{Margaret M. Coad,}
Department of Mechanical Engineering, Stanford University, Stanford, California, USA. E-mail: mmcoad@stanford.edu.
\end{IEEEbiographynophoto}
\vskip -2.5\baselineskip plus -1fil
\begin{IEEEbiographynophoto}{Laura H. Blumenschein,}
Department of Mechanical Engineering, Stanford University, Stanford, California, USA. E-mail: lblumens@stanford.edu.
\end{IEEEbiographynophoto}
\vskip -2.5\baselineskip plus -1fil
\begin{IEEEbiographynophoto}{Sadie Cutler,}
Department of Mechanical and Aerospace Engineering, Cornell University, Ithaca, New York, USA. E-mail: sc3236@cornell.edu.
\end{IEEEbiographynophoto}
\vskip -2.5\baselineskip plus -1fil
\begin{IEEEbiographynophoto}{Javier A. Reyna Zepeda,}
Department of Mechanical Engineering, Stanford University, Stanford, California, USA. E-mail: jreynaza@gmail.com.
\end{IEEEbiographynophoto}
\vskip -2.5\baselineskip plus -1fil
\begin{IEEEbiographynophoto}{Nicholas D. Naclerio,}
Department of Mechanical Engineering, University of California Santa Barbara, Santa Barbara, California, USA. E-mail: nnaclerio@ucsb.edu.
\end{IEEEbiographynophoto}
\vskip -2.5\baselineskip plus -1fil
\begin{IEEEbiographynophoto}{Haitham El-Hussieny,}
Electrical Engineering Department, Faculty of Engineering (Shoubra), Benha University, Egypt. E-mail: haitham.elhussieny@feng.bu.edu.eg.
\end{IEEEbiographynophoto}
\vskip -2.5\baselineskip plus -1fil
\begin{IEEEbiographynophoto}{Usman Mehmood,}
School of Mechanical Engineering, KOREATECH, Cheonan City, Chungnam Province, Republic of Korea. E-mail: usman.mehmood18@gmail.com.
\end{IEEEbiographynophoto}
\vskip -2.5\baselineskip plus -1fil
\begin{IEEEbiographynophoto}{Jee-Hwan Ryu,}
Department of Civil and Environmental Engineering, Korea Advanced Institute of Science and Technology, Daejeon, Republic of Korea. E-mail: jhryu@kaist.ac.kr.
\end{IEEEbiographynophoto}
\vskip -2.5\baselineskip plus -1fil
\begin{IEEEbiographynophoto}{Elliot W. Hawkes,}
Department of Mechanical Engineering, University of California Santa Barbara, Santa Barbara, California, USA. E-mail: ewhawkes@ucsb.edu.
\end{IEEEbiographynophoto}
\vskip -2.5\baselineskip plus -1fil
\begin{IEEEbiographynophoto}{Allison M. Okamura,}
Department of Mechanical Engineering, Stanford University, Stanford, California, USA. E-mail: aokamura@stanford.edu.
\end{IEEEbiographynophoto}

\end{document}